\documentclass{article}
\usepackage{arxiv}

\usepackage[utf8]{inputenc} 
\usepackage[T1]{fontenc}    
\usepackage{hyperref}       
\usepackage{url}            
\usepackage{booktabs}       
\usepackage{amsfonts}       
\usepackage{nicefrac}       
\usepackage{microtype}      
\usepackage{lipsum}		
\usepackage{graphicx}
\graphicspath{ {./figures/} }
\usepackage{doi}
\usepackage{amsmath}
\usepackage{float}

\title{Temporal-Spatial Processing of Event Camera Data via Delay-Loop Reservoir Neural Network }

\begin{document}
\hypersetup{
pdftitle={Temporal-Spatial Processing of Event Camera Data via Delay-Loop Reservoir Neural Network},
pdfsubject={ML, CV},
pdfauthor={Richard Lau, Anthony Tylan-Tyler, Lihan Yao, Rey de Castro Roberto, Robert Taylor, Isaiah Jones},
pdfkeywords={Temporal-Spatial Conjecture, Video Markov Model, Mutual Information, Mutual Information Neural Estimate, Delay Loop Reservoir, Event Camera, Edge Application.},
}
\author{{Richard Lau (\texttt{clau@peratonlabs.com})}, {Anthony Tylan-Tyler (\texttt{anthony.tylan-tyler@peratonlabs.com})}, \\
{Lihan Yao (\texttt{lihan.yao@peratonlabs.com})}, 
{Rey de Castro Roberto (\texttt{rreydecastro@peratonlabs.com})}, \\
{Robert Taylor (\texttt{robert.m.taylor@peratonlabs.com})}, 
{Isaiah Jones (\texttt{isaiah.l.jones@peratonlabs.com})}
}

\date{Peraton Labs \\ 331 Newman Springs Road, Red Bank NJ, 07701, USA}

\renewcommand{\headeright}{R. Lau et al.}
\renewcommand{\undertitle}{}


\maketitle

\begin{abstract}
This paper describes a temporal-spatial model for video processing with special applications to processing event camera videos. We propose to study a conjecture motivated by our previous study of video processing with delay loop reservoir (DLR) neural network \cite{HyDDENN, scaled}, which we call \textbf{\emph{Temporal-Spatial Conjecture}} (TSC). The TSC postulates that there is significant information content carried in the temporal representation of a video signal and that machine learning algorithms would benefit from separate optimization of the spatial and temporal components for intelligent processing. To verify or refute the TSC, we propose a \textbf{\emph{Visual Markov Model}} (VMM) which decompose the video into spatial and temporal components and estimate the mutual information (MI) of these components. Since computation of video mutual information is complex and time consuming, we use a Mutual Information Neural Network\cite{mutualinfo} to estimate the bounds of the mutual information.  Our result shows that the temporal component carries significant MI compared to that of the spatial component. This finding has often been overlooked in neural network literature. In this paper, we will exploit this new finding to guide our design of a \textbf{\emph{delay-loop reservoir}} neural network for event camera classification, which results in a \textbf{\emph{18\% improvement}} on classification accuracy. 
\end{abstract}


\section{Introduction}
While there are many Artificial Intelligence/Machine Learning (AI/ML) techniques in literature that are used in practice for classification and prediction of video signals, most employ techniques that are handcrafted by experts in the field, often without theoretical justifications \cite{manifoldlearning, exploiting}. In this paper, which is based upon work supported by the Defense Advanced Research Projects Agency (DARPA) on “Reservoir-based Event Camera for Intelligent Processing on the Edge (RECIPE)”, we decompose a video signal into temporal and spatial component and postulate a \textit{Temporal-Spatial Conjecture} (TSC) that suggests benefits by separating temporal and spatial processing to reduce training time and complexity. To validate or refute the TSC, we propose a \textit{Video Markov Model} (VMM) that captures inter-dependence of key components of the visual signal. The VMM quantifies the dependence by measuring by their \textit{mutual information} (MI). Results from the VMM provide insight to help explain why certain approaches work and possibly guidance for the construction of improved AI/ML algorithms. The TSC study was motivated by a prior result from our previous work in HyDDENN \cite{HyDDENN, scaled}, where we had found that seperate temporal/spatial processing in a drone/bird classifier leads to an improvement of Signal-to-Noise ratio compared to a classifier that processes the combined temporal-spatial information. In this paper, we aim to quantify such advantage. 

To demonstrate the TSC, we apply the lessons learned from the VMM to process event camera videos. Recent years have seen growing research efforts in “Event Cameras”, which provides a new way of capturing visual sensing in the emerging field of neuromorphic engineering. Event cameras are fundamentally different from traditional visual cameras, and has created a paradigm shift in how visual information is acquired. A traditional camera samples the scene periodically (e.g. 30 frames/sec) and captures light intensity on all the pixels. In contrast, an event camera records only differential pixel intensity information upon scene changes. This represents a fundamentally different method of capturing visual information and leads to the following benefits: 1) the sensed data is sparse and significantly smaller and thus reduces storage and transport requirements by a factor of 1000×; 2) ultra-high temporal resolution and low latency (in microseconds); and 3) very high dynamic range (140dB), and 4) low power consumption. These characteristics have great impact on the next generation of sensors deployed in both military and commercial systems, including self-driving automobiles, high-speed Unmanned Aerial Vehicles (UAV), robotics, ultra-high-speed data transfer, storage reduction, and smart surveillance. 

To qualify the potential improvement, we will apply the TSC-inspired architecture to process event video via a Delay-Loop Reservoir Neural Network (DLR) for classifying and predicting spike signals from event-based cameras. Advantages of DLR for edge networks have been studied and reported in \cite{scaled}. In this paper, we will demonstrate how TSC is used in improving DLR for processing event videos. 

This paper is organized as follows. In Section 2, we describe the TSC and its motivation, followed by defining the VMM structure. Section 2 also includes a description of the DVS event dataset that we use to perform the study. Section 3 goes into detailed description of the Mutual Information Neural Estimate approach for estimating the Mutual Information for the VMM. It then describes the results when the VMM is applied to the DVS data. Motivated by the TSC result and insight, Section 4 applies the Delay-Loop Reservoir (DLR) for classification of the DVS data. It compares performance results between a direct DLR vs. a modified DLR as inspired by the TSC and explains how lessons learned from the TSC helps to design a better architecture for the DLR. Section 5 discusses future research in extension of the study and implications of the TSC. Section 6 gives the final conclusion. 

\vspace{-10truemm}
\section{Temporal Spatial Conjecture}
\vspace{-10truemm}
\label{sec:Temporal Spatial Conjecture}

The study of the TSC is motivated by understanding the fundamental theory behind processing video signals. Video processing algorithms often process the video signal as a 3-dimensional tensor of space and time, which requires large amount of memory and computational resource. Since a video signal is made up of consecutive video frames, each of which are static images, the spatial content of nearby frames is often highly correlated or even repetitive. However, most current AI/ML techniques used for classification and prediction of video signals do not exploit this correlation. While well-designed AI/ML algorithms, often handcrafted by experts in the field, can extract this correlation to achieve the goal, the algorithms are excessively computationally demanding, which becomes a significant problem for efficient edge applications.

The current study asks the following question: Are there efficient ways to break up the video signal before processing which may lead to more efficient resource usage? We start with postulating a conjecture, called Temporal-Spatial Conjecture (TSC), which states that separating the temporal and spatial processing is beneficial in visual processing to reduce training time and complexity. The TSC was motivated by a prior result from our previous work in HyDDENN \cite{HyDDENN, scaled}, where we had found that separating temporal/spatial processing in a drone/bird classifier leads to an improvement of Signal-to-Noise ratio of 4dB compared to a classifier that processes the combined temporal-spatial information. In this paper, we will build an analytic tool to quantify such advantage. 
\vspace{-10truemm}
\subsection{Video Model}
To analyze the TSC, we build a Video Markov Model (VMM) for analysis of information content of the components of the video signal. We first represent an event video ($V$) as a set of contiguous event frames, 
$F_t, $for $  t=1, ..., D$
 are the time samples. The frames are sampled in time, which can be non-uniform. For notational simplicity we assume uniform sampling in the following. Thus, we write, 
\begin{equation*}
V = {F_t}               \hspace{150pt}   t = 1, ..., D
\end{equation*}
The VMM is created via two procedures. First it decomposes the event-based video signal ($V$) into a spatial component ($S$) and a temporal component ($T$). The decomposition should satisfy two criteria: 
\begin{enumerate}
	\item Each component $S$, $T$, is a subset of the events of the original signal $V$.
	\item The union of $S$ and $T$, carries information that approaches that of the entire signal. Such information is measured with respect to an AI/ML application such as predicting or classifying the label of the video.
\end{enumerate}

\begin{center}
    \includegraphics[width=200pt]{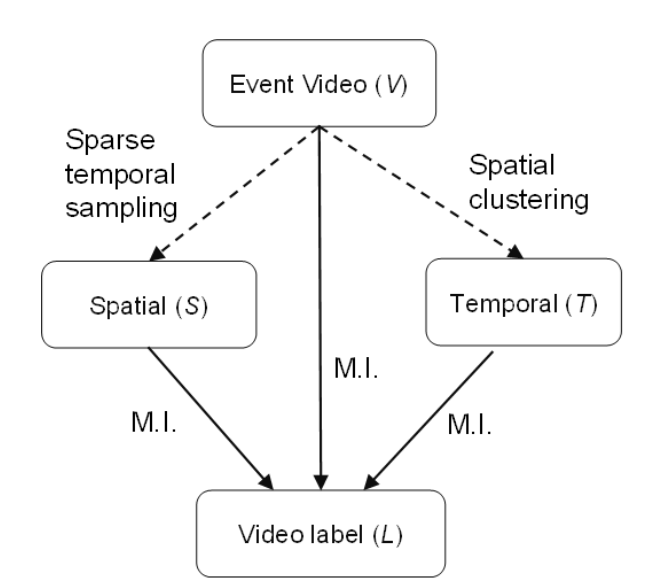} \\
	Figure 1: Video Markov Model
	\label{fig:fig1}
\end{center}
 As shown in Figure 1, the spatial component $S$ is defined to be a set of sampled event frames such that the frames have negligible correlation among them. Thus, we write
 \begin{equation*}
S = \{F_{t_u}\}               \hspace{100pt}   t_u = \text{time index of uncorrelated frames}
\end{equation*}
There are different ways to obtain the spatial component. One way is to randomly shuffle the event frames so that they are uncorrelated. Another method is to sparsely sample the event frames such that there is little correlation between the samples. Note that both methods satisfy criteria 1 given above. However, the later method is usually preferred since it reduces processing and storage complexity.  

While the spatial component modifies and reduces temporal content, the temporal component retains temporal information but allows reduction of the spatial information. There are numerous ways that $T$ can be defined to satisfy the VMM criteria. We propose to define $T$ as clusters of events in contiguous frames. Thus, we write
\begin{equation*}
 T = \{C_t\}               \hspace{150pt}   t = 1, ..., D
\end{equation*}
where $C_ts$ are clusters of the events at frame of time $t$. This definition does not place stringent requirements on the structure of the clusters but requires that the temporal information be unchanged from that of the original video signal.  
The second part of the VMM is to compute or learn the mutual information for 3 configurations as shown in Figure 2. We describe in more detailed in the following section on the computation and estimation of these MIs {\emph{I}. The amount of MI carried in these models provide answer to the TSC and hints on how to design efficient AI/ML architectures. 
\begin{center}
    \includegraphics[width=300pt]{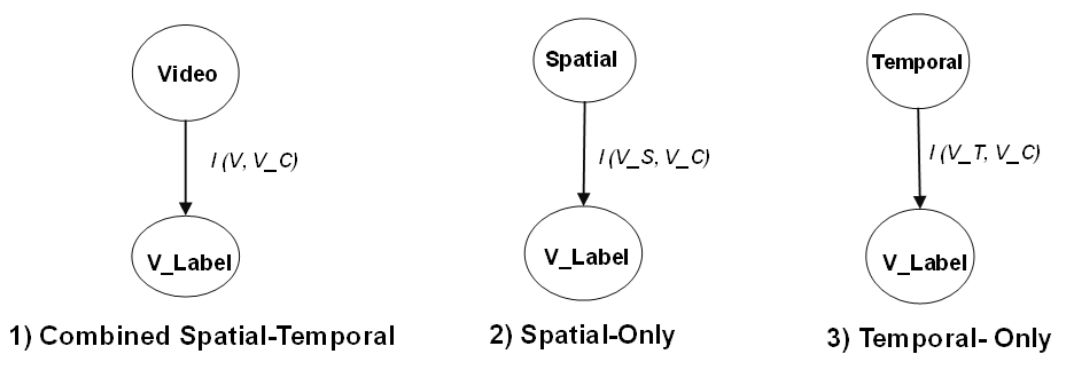} \\
	Figure 2: VMM sub structure diagrams for study of TSC 
	\label{fig:fig2}
\end{center}
\subsection{Event Data}
For testing of the VMM and testing of subsequent even data classification, we will use Dynamic Vision Sensor (DVS) dataset \cite{dvsdataset}. DVS is a low-power, real-time, event-based hand gesture recognition system. It is the first implemented entirely on event-based hardware. DVS captures changes in pixel brightness, transmitting data only when a change is detected, which significantly reduces power consumption compared to traditional frame-based cameras. The DVS dataset comprises of 11 hand gesture categories from 29 subjects under 3 illumination conditions (see Figure 3).  
\begin{figure}[!htbp]
\centering
    \includegraphics[width=400pt]{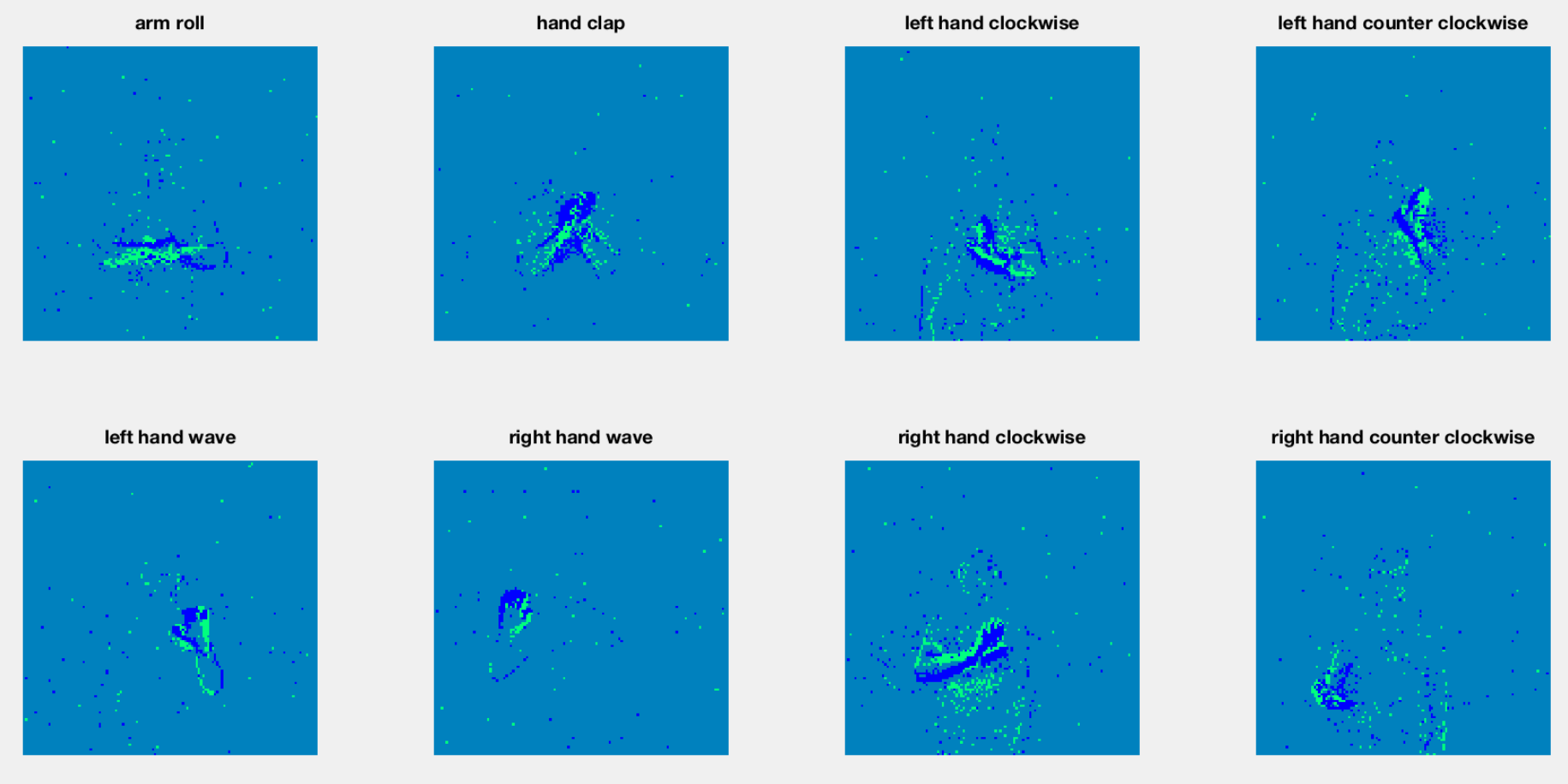} \\
	Figure 3: Examples from 8 of the 11 classes in the DVS dataset
	\label{fig:fig3}
\end{figure}
\begin{figure}[!htbp]
    \centering
    \includegraphics[width=300pt]{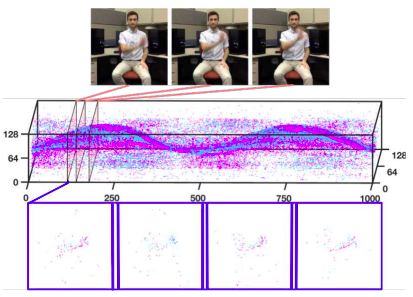} \\
	Figure 4: VMM RGB captures of gestures, the corresponding event stream, and the aggregation of events into frames.
	\label{fig:fig4}
\end{figure}

To process the data, the event stream is first partitioned in time as shown in Figure 4, where events in a partition, i.e. a time window, is aggregated into one frame. In the DVS gesture dataset, each frame collects events over a time window of 30 milliseconds. By sequencing 200 consecutive frames, an observation represents $\sim$6 seconds in time.

\section{Mutual Information Extraction}
\label{sec:Mutual Information Extraction}
Solving the TSC requires computation of Mutual Information, $I$, which is a difficult task due to the large size of the data. This section describes how we use the Mutual Information Neural Estimate (MINE) \cite{mutualinfo} method to estimate $I$. 

\subsection{Mutual Information}
The mutual information between two variables is a measure of how dependent they are on each other. For a pair of variables where one is known, the mutual information quantifies how much is known about the unknown variable from the known variable. If the mutual information is high, then the unknown variable is well constrained by the known variable. If the mutual information is low, then the unknown variable is relatively unconstrained by the known variable. Applying this to the TSC, we can use the mutual information to see how well the separate spatial and temporal information can be associated with the video label. 

The mutual information is defined as the difference between the entropy of a single variable, $H)X = \mathbb{E}_p [log p(x)]$, where $\mathbb{E}_P[x]$ is the expectation value of $x$ for the distribution $P$, and the joint entropy of both, $H(X | Y):$
\begin{equation*}
I(X,Y) = H(X) - H(X | Y)
\end{equation*}
The mutual information between the two variables is constrained from above by $I(X,Y) \le min⁡(H(X),H(Y))$ This constraint arises from $I(X,Y)=I(Y,X)$ and that $min(H(X|Y)) = 0$ when $X$ is fully determined by $Y$. Thus the maximum mutual information occurs when one variable is fuly determined by the other and is then constrained by which variable has the lowest entropy. This constraint will allow us to determine when the mutual information is 'high' as well as compare $I(V_S, V_C)$ and $I(V_T, V_C)$ where $V_S$ is the spacial information of the video, $V_T$ is the temporal information of the video, and $V_C$ is the video class label.

\subsection{Mutual Information Neural Estimate}
Determining the mutual information $I(X,Y)$ can be done when the joint distribution of the variables is known. In our case, these distributions are not known and, further complicating matters, the video data is very high dimensional. For the individual frames, for example, the probability distribution is over a 32768 dimensional space. Due to its high dimensionality, the sample frames are likely to lie very far from each other, making the estimation of the whole distribution unfeasible. In order to overcome this limitation, we use the Mutual Information Neural Estimation (MINE) \cite{mutualinfo} to estimate the mutual information between pairs of variables. 
\begin{figure}[!htbp]
    \centering
    \includegraphics[width=300pt]{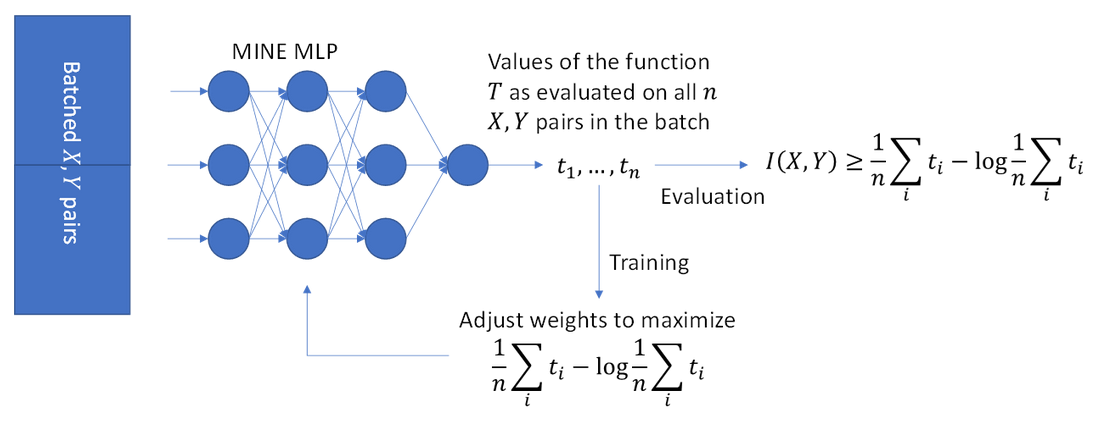} \\
	Figure 5: A diagrammatic representation of the MINE algorithm
	\label{fig:fig5}
\end{figure}

MINE is a neural network which estimates the mutual information using an approximate calculation of the Kullback-Leiber (KL) divergence. To begin, the mutual information can be rewritten in terms of the KL diverence between the joint distribution $P(X|Y)$ and the product of the marginal distribution $P_X \oplus P_Y$
\begin{equation*}
I(X,Y) = D_{KL}(P(X|Y)   \hspace{2pt}||  \hspace{2pt}  P(X) \oplus P(Y))
\end{equation*}
The KL divergence between two distributions is defined as 
\begin{equation*}
I(X,Y) = D_{KL}(P||Q) = \mathbb{E}_P[log(\frac{P}{Q})]
\end{equation*}
$D_{KL}$ can be estimated using the Donsker-Varadhan representation as
\begin{equation*}
D_{KL}(P||Q) \geq \sup_{T' \in \mathcal{F}} \mathbb{E}_P[T'] - log(\mathbb{E}_Q[e^{T' - 1}])
\end{equation*}
where $T'$ is a function and $\mathcal{F}$ is a subset of functions. Using this inequality, the neural network shown in Figure 2 attempts to learn the function $T'$ which maximizes the Donsker-Varadhan representation of the KL divergence.
\begin{figure}[!htbp]
    \centering
    \includegraphics[width=300pt]{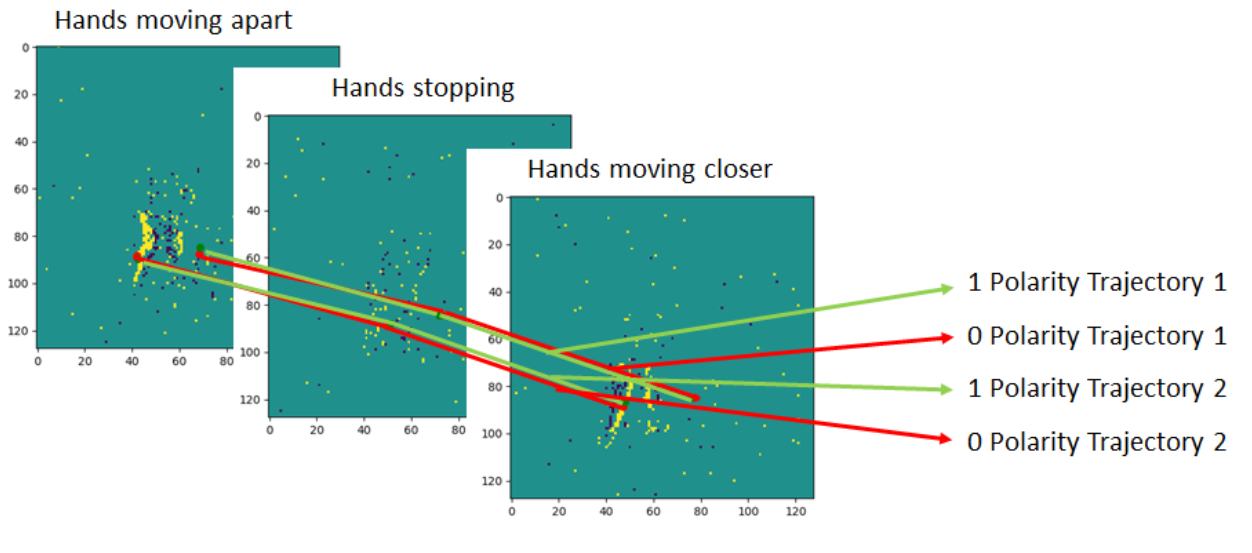} \\
	Figure 6: An example of the trajectories making up $V_T$ from a video of the ‘clapping hands’ class of the DVS128 Gesture dataset
	\label{fig:fig6}
\end{figure}
Applying MINE to video data requires us to rigorously define the video spatial ($V_S$) and temporal ($V_T$) information. We choose to represent the $V_S$ as a segment of 5 full resolution images covering 0.5s of video (a flattened 5x2x128x128 tensor). $V_T$, is represented as a flattened 8x200 matrix of the trajectories of 4 centroids defined by $k$-means clustering on the event stream which has been sliced along time into segments of 0.027s (see Figure 6). 

Using these definitions, the MINE can be used to calculate the entropy of each variable as $I(X, X) = H(X)$. This gives us $H(V_S) > 24$, and $H(V_C) \geq 2.49$. As the MINE estimate is established to be tight, we use $H(V_T)$ as the maximum value for both $I(V_S, V_C)$ and $I(V_T, V_C)$. Thus we can directly compare the values of $I(V_S, V_C)$ and $I(V_T, V_C)$ to understand the relative information content of each. Next, we can use MINE to estimate $I(V_S, V_C) \geq 2.13$ (85\% of maximum) and $I(V_T, V_C) \geq 2.31$ (92\% of maximum)

\subsection{Implication of Mutual Information on TSC}
These results suggest that for event camera data, the bulk of the information for labelling the video data is in the temporal information. As we highlighted before, it is difficult to separate the spatial and temporal information. For example, $V_T$ is composed of the spatial location of cluster centroids over time which still includes spatial information. Similarly, our definition of $V_S$ covers 0.5s of events over 5 frames (0.1s of events compiled into each frame), which includes changes over time. If, instead, a single frame compiled from 0.1s of events is used for $V_S$, the mutual information drops to 1.26 (50\% of maximum), suggesting that the spatial information alone contains significantly less information necessary for labelling the videos. 

An additional complication of the spatial information is its high dimensionality. At a resolution of 2x128x128, the resulting vector has 32768 components. In order to simplify the spatial information while retaining the temporal information which appears to contain the bulk of the video information, we propose to down sample the spatial resolution to 2x8x8 over 200 frames, giving vectors of 25600 a 21\% reduction in size to cover the whole video compared to a single frame while still retaining the bulk of the spatial information. Labelling this the full video information, $V$, we find that $I(V,V_C) \geq 2.44$ (98\% of maximum). Thus, by simplifying the spatial information through down sampling while retaining all of the temporal information, we are able to improve our result by 6\%. This result is significant as it motivates us to design a corresponding DLR to take advange of the insight learned. This will be verified in an DLR application in the following section.
\section{DLR Classification of Event Camera Data}
This section will illustrate how the insight gained from the TSC study motivate an improvement on the Delay Loop Reservoir (DLR) algorithm for event camera processing. 
\subsection{DLR Algorithm}
\begin{figure}[!htbp]
    \centering
    \includegraphics[width=300pt]{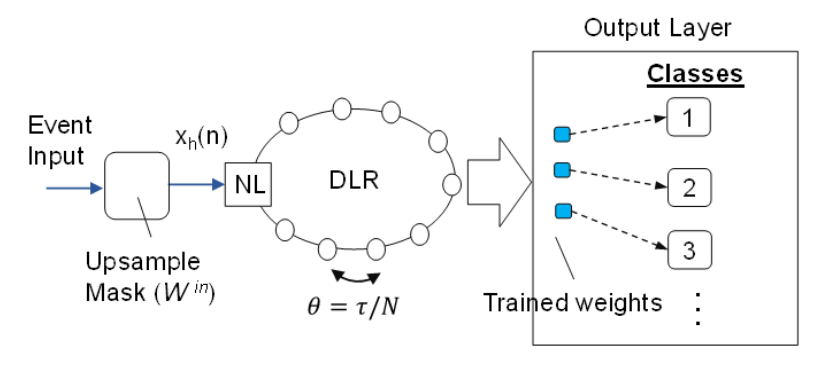} \\
	Figure 7: Delay-Loop Reservoir structure
	\label{fig:fig7}
\end{figure}
As shown in Figure 7, the Delay Loop Reservoir (DLR) \cite{scaled} is a nonlinear, high dimensional model within the reservoir computing framework. The DLR operates on serialized stream data by sequentially upsampling each sample into high dimensions, via random weights. The resulting high dimensional representation is then processed by the core delay loop, which serves two functions: 1) inject nonlinearity using hyperbolic tangent; and 2) allows processing temporal processing via a leaky factor which combine past temporal data with current data. The leaky factor thus controls the degree of “memory” of the DLR, and can be used to select frequency components for processing. After loop processing, the data is classified by a linear classifier such as a ridge regressor. Note that the entire DLR has only training at the ridge regressor stage and the delay loop itself does not required training. This is a main difference between reservoir based NN and other NN’s. The simplicity of the DLR architecture leads to practical implementation in hardware, which has been demonstrated in FPGA implementation as reported in \cite{scaled}.

\subsection{Initial Result DLR Algorithm}

We first apply the DLR on the 11-class classification task of the DVS dataset with tuning parameters including the leaky factor, loop dimensionality, and ridge regression regularization. Our initial result showed an accuracy of 76\%. We noted that this sub-par performance can be attributed to an overfitting problem. Overfitting occurs when a model learns the training data too well, including its noise and outliers, and therefore performs poorly on unseen data. This is especially prevalent for event camera data, where noise among events in an event stream may become a ‘crutch’ for the model. As an illustration, Figure 8 shows the spurious noisy events that cause overfitting. The red blob in the event stream has been memorized by the model during training.  Whenever the spurious event appears, the model predicts class $i$. However, in the test set, an event stream of class $i$ does not contain this specific event.  

\begin{figure}[!htbp]
    \centering
    \includegraphics[width=400pt]{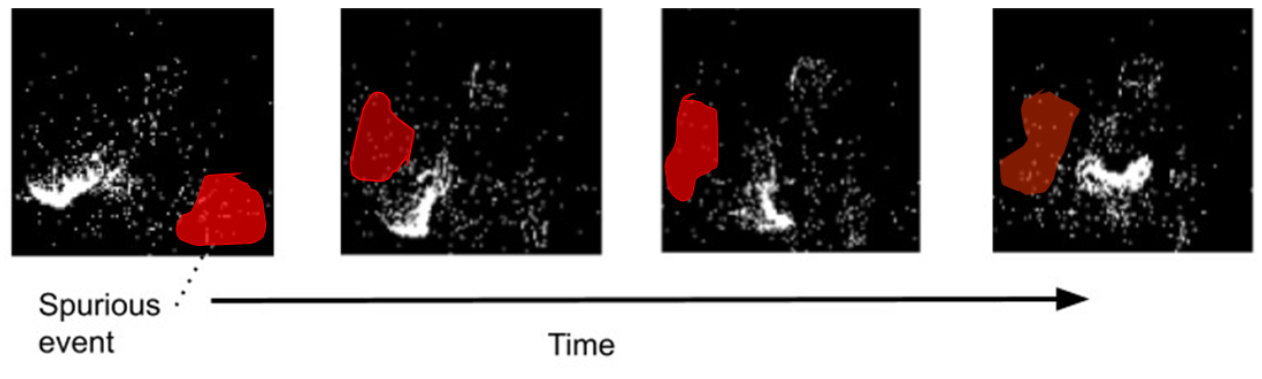} \\
	Figure 8: Example of spurious events in the event stream 
	\label{fig:fig8}
\end{figure}

\subsection{TSC-inspired modification of the DLR Algorithm}
As illustrated above, the problem of overfitting in event classification is that the DLR tries to classify using the noisy data. It does a good job for the training data and indeed this is what had found. During training, the DLR performs close to 100\% accuracy while its accuracy falls to 76

In traditional NN, methods to mitigate overfitting in neural networks include reducing the number of weights in the NN or generation of new training data by data augmentation. These would not work in DLR as the loop does not have training parameter and regularization in the ridge regressor was already used. Data augmentation is also not desirable as it adds complexity to the model.  
\begin{figure}[!htbp]
    \centering
    \includegraphics[width=300pt]{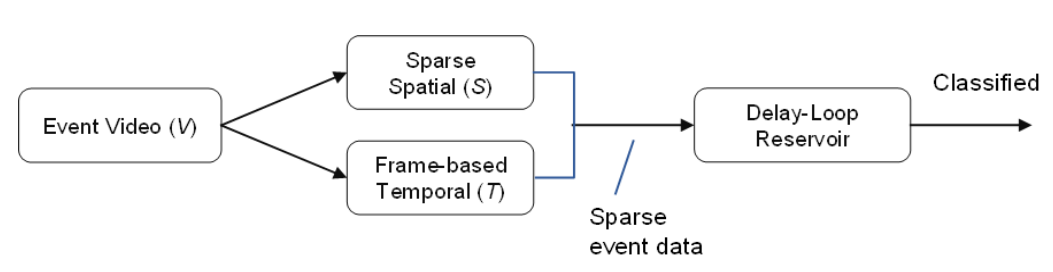} \\
	Figure 9: TSC-inspired preprocessing for DLR improvement
	\label{fig:fig9}
\end{figure}
Instead, we will apply a data reduction module to the DLR as inspired by the TSC result. This modification is shown in Figure 9. The Sparse Spatial module suggests lowering the sampling rate of the events while the Temporal module captures frames from the event. To tackle the overfitting problem, we will substantially reduce the spatial sampling rate, but retaining sufficient temporal resolution, which is consistent with the TSC result that the temporal component carries more information for label classification. Using this modification, the new DLR architecture is shown in Figure 10. 
\begin{figure}[!htbp]
    \centering
    \includegraphics[width=300pt]{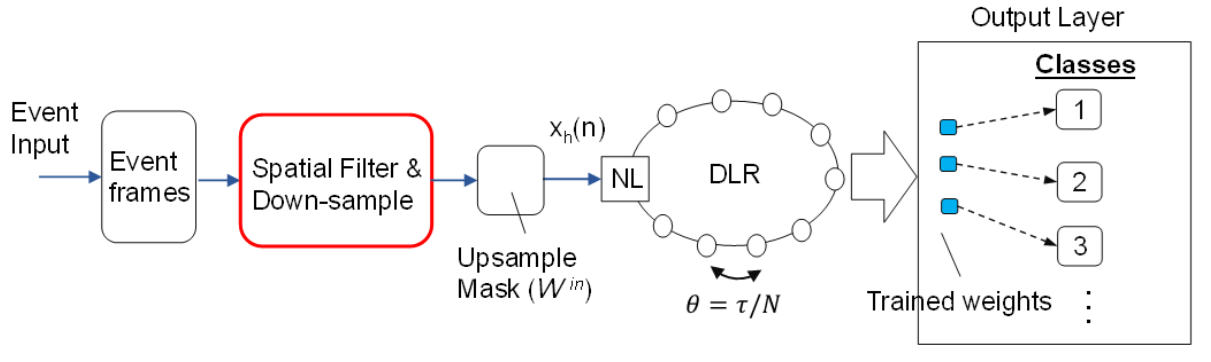} \\
	Figure 10: Modified DLR Architecture
	\label{fig:fig10}
\end{figure}
The main addition of Figure 10 is the Spatial Filter and Down-sample module. We apply a low-pass filter to remove high-frequency noise that could lead to aliasing artifacts before subsampling. The subsampling process reduces the image size by keeping every $n$-th pixel in each row and column, and discarding the rest. The procedure $g$ is applied to pixel coordinate $(i, j)$, given by
\begin{equation*}
g(i, j) = \sum_{k, l}^n f(k, l)h(i-\frac{k}{r}, j - \frac{l}{r})
\end{equation*}
\begin{doublespace}
where $f(k,l)$ is the smoothing filter, $h(i-\frac{k}{r}, j - \frac{l}{r})$ subsamples the blurred image to dimensions r by r. Figure 11 shows effect of the sampling process for various subsampling rates.
\end{doublespace}
\begin{figure}[!htbp]
    \centering
    \includegraphics[width=300pt]{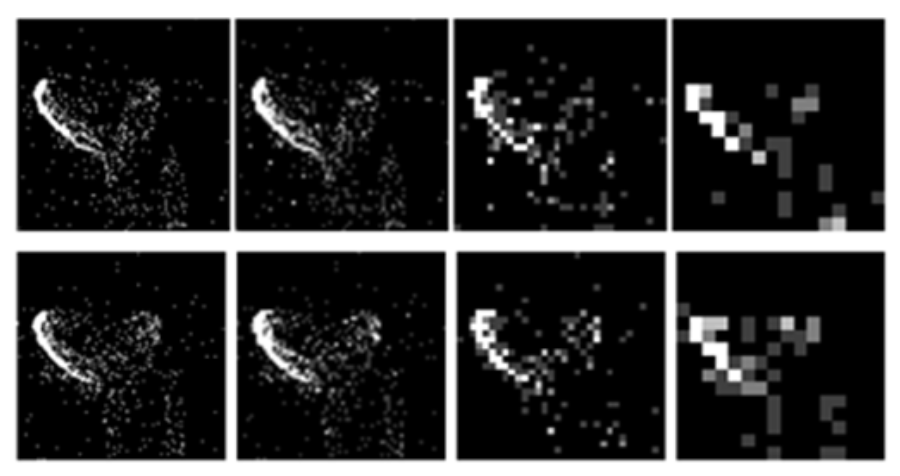} \\
	Figure 11: Image of subsampling effect
	\label{fig:fig11}
\end{figure}
\begin{figure}[!htbp]
    \centering
    \includegraphics[width=300pt]{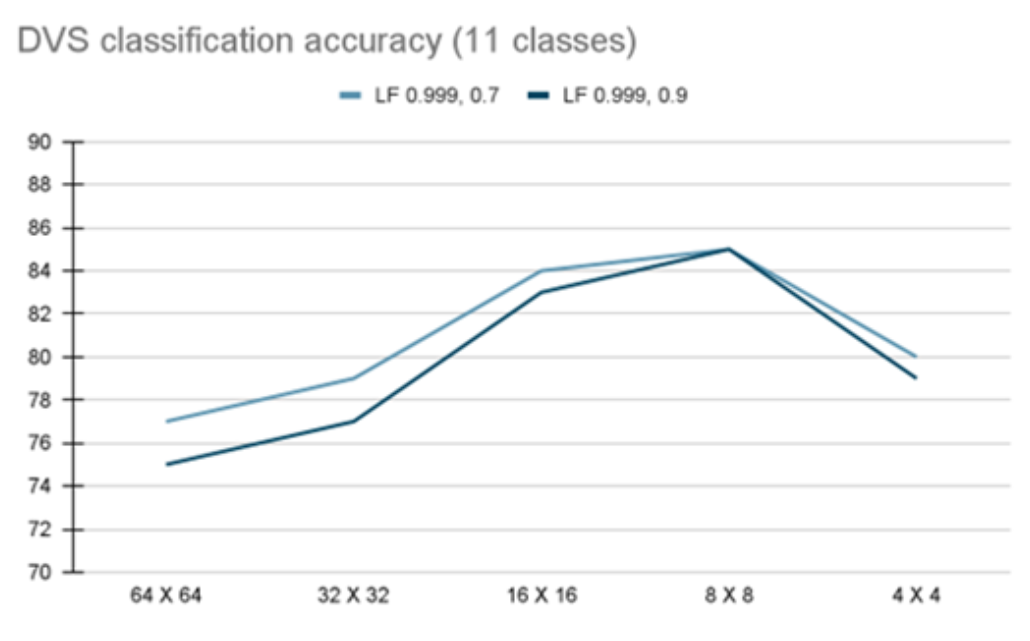} \\
	Figure 12: DLR Accuracy with respect to different frame sub-sampling
	\label{fig:fig12}
\end{figure}

Figure 12 shows the performance improvement of the modified DLR architecture. It shows that the largest improvement is achieved by downsampling the event frame to an 8 × 8 event pixel frames. This result cannot be adequately explained by traditional methods to tackle overfitting in machine learning, as traditional NN methods focus on spatial learning. However, this result can be fully explained from the viewpoint of the TSC as temporal component carries more information. Since noise is most prominent in the spatial component, we can substantially reduce the frame size and thus reducing overfitting to a minimum without affecting the temporal component.

Using the modified DLR architecture, with fine tuning of other DLR parameters including leaky factor, loop dimensionality, we were able to achieve an accuracy of 89.6\% for the 11-class DVS data classification. This is in comparison with an accuracy of 96\% \cite{sabater} when attention neural networks are used. Thus, the DLR has achieved on-par performance with State of the Art Attention Neural Network but with substantial reduction in training time and computation complexity as measured by Multiply-and-Accumulate units \cite{scaled}.

\section{Future Work}
This paper is motivated by exploring processing of the temporal component of a video signal as this area of research seems to have not been the focus in machine learning algorithm development. The proposed VMM validates an important result for applying DLR to event data by optimizing the spatial component. We have not investigated optimization with respect to the temporal domain. For example, what would be an optimum frame structure? And what would be a joint spatial-temporal optimum structure? Also, we would also like to study the impact of TSC on other applications such as object segmentation and object recognition.

\section{Conclusion}
This paper describes a temporal-spatial model for video processing with special applications to processing event camera videos. We first describe the Temporal-Spatial Conjecture which postulates that there is significant information content carried in the temporal representation of a video signal. TSC also advocates that machine learning algorithms would benefit from separate optimization of the spatial and temporal components for intelligent processing. We described the Visual Markov Model which decomposed videos into spatial and temporal components and estimate the mutual information of these components with respect to machine learning functions. The VMM study found that temporal component carries more information than its spatial counterpart. The finding is important, which was exploited to help design a modification to the delay-loop reservoir algorithm for classification of DVS event data. The modified DLR algorithm tackles the overfitting problem successfully and achieved 18\% improvement on classification accuracy, thereby validating the TSC. 

\section{Acknowledgement}
This paper is based upon work supported by the DARPA MTO contract N6600122C4025 on “Reservoir-based Event Camera for Intelligent Processing on the Edge (RECIPE)". We thank Dr. Whitney Mason and Dr. Timothy Klausutis, Program Managers, for their vision on the TSC and DLR and guidance of the program, Greg Jones for his guidance and technical discussions on the program, and Justin Mauger (CIV USN NIWC PACIFIC CA (USA)) for many technical discussions. The views, opinions and/or findings expressed are those of the authors and should not be interpreted as representing the official views or policies of DARPA or the U.S. Government.

\bibliographystyle{unsrtnat}


\end{document}